\documentclass[lettersize,journal, twoside]{IEEEtran}
\usepackage{amsmath,amsfonts}
\usepackage{algorithmic}
\usepackage{algorithm}
\usepackage{array}
\usepackage{textcomp}
\usepackage{stfloats}
\usepackage{url}
\usepackage{verbatim}
\usepackage{cite}
\usepackage{graphicx}

\usepackage{graphics}
\usepackage{epsfig} 
\usepackage{subcaption}
\usepackage{setspace}
\usepackage{bm}
\usepackage{upgreek}
\usepackage[labelsep=period]{caption} 
\usepackage[british]{babel} 

\begin{document}

\title{3D Uncertain Implicit Surface Mapping using GMM and GP}

\author{Qianqian Zou and Monika Sester
\thanks{
This work was supported by the German Research Foundation (DFG) as part of the Research Training Group i.c.sens. (corresponding author: Qianqian Zou)}
\thanks{The authors are with the Institut of Cartography and Geo-informatics in Leibniz University Hannover, Appelstraße 9a 30167 Hannover, Germany (email: qianqian.zou@ikg.uni-hannover.de; monika.sester@ikg.uni-hannover.de).}
}

\markboth{}
{ZOU \MakeLowercase{\textit{et al.}}: 3D Uncertain Implicit Surface Mapping using GMM and GP}


\maketitle

\begin{abstract} 
    In this study, we address the challenge of constructing continuous three-dimensional (3D) models that accurately represent uncertain surfaces, derived from noisy and incomplete LiDAR scanning data. Building upon our prior work, which utilized the Gaussian Process (GP) and Gaussian Mixture Model (GMM) for structured building models, we introduce a more generalized approach tailored for complex surfaces in urban scenes, where GMM Regression and GP with derivative observations are applied.  
    A Hierarchical GMM (HGMM) is employed to optimize the number of GMM components and speed up the GMM training. With the prior map obtained from HGMM, GP inference is followed for the refinement of the final map. Our approach models the implicit surface of the geo-object and enables the inference of the regions that are not completely covered by measurements. The integration of GMM and GP yields well-calibrated uncertainties alongside the surface model, enhancing both accuracy and reliability. The proposed method is evaluated on real data collected by a mobile mapping system. Compared to the performance in mapping accuracy and uncertainty quantification of other state-of-the-art methods, the proposed method achieves lower RMSEs, higher log-likelihood values and lower computational costs for the evaluated datasets.
\end{abstract}


\section{Introduction}
\IEEEPARstart{A}{CCURACY} and the possibility to detect gross errors are both critical for mapping in unknown unstructured urban environments, as they guarantee uncertainty-aware localization, secure navigation and optimal path-planning of autonomous systems, especially with noisy and incomplete measurements. Some popular mapping frameworks \cite{octomap, nerf, gs} emphasize accuracy and efficiency while the uncertainty of mapping outputs is not sufficiently addressed.  
Since large errors can also appear in a generally accurate model due to data errors and incompleteness, it is necessary to quantify the uncertainty of the mapping results as well.

The characteristics of the environment representation greatly affect the downstream uncertainty-aware tasks. Traditional occupancy maps \cite{octomap} represent the space as discrete structured cells and the occupancy of each cell is modelled independently by ray-casting techniques. While they are widely used in localization and collision avoidance, the inconsistency of independent structures poses the challenge for the incomplete, noisy and sparse LiDAR measurements in urban areas. As an alternative, continuous implicit models \cite{gpis0, gp-implicit-map, gpis1, sdf, loggpis2, kigp, gpis2} show their advantages in solving this problem. For example, signed-distance fields (SDF) have been suggested to improve the accuracy and robustness of surface modelling \cite{sdf}. 

In this work, we aim to build continuous 3D surface models of outdoor urban areas using implicit signed-distance functions, together with uncertainty measures. Probabilistic inferences offer the possibility to continuously model environments with well-calibrated uncertainties. Therefore, Gaussian Mixture Model (GMM) and Gaussian Process (GP) inferences are employed and combined in our approach to estimate the distance fields in large-scale scenes, especially with urban manmade structures, based on LiDAR point clouds. Additionally, we focus on the uncertainty of mapping outputs and investigate if a proper uncertainty measure can provide a good insight into the mapping process. That means the map quality is correctly quantified and poor map estimates can be avoided accordingly. The existing methods of continuous mapping currently do not sufficiently address this uncertainty issue. 

To generate continuous maps, GP-mapping \cite{gpom} introduces the spatial correlation to the neighbouring points. In GP occupancy maps \cite{gpom, gpmap, gpom_fast}, the occupancy states are predicted by GPs and "squashed" into probability in the range of [0, 1] by a logistic regression, which is robust with sparse data. To reduce the computation time, approximation methods like sparse kernel\cite{sparsekernel} and data partition\cite{gpmap} are used in large-scale scenes.  
Alternatively, Hilbert maps \cite{hm1} and Bayesian kernel inference \cite{bgkl} search for a comparable accuracy as GP, but with faster prediction time. However, the quality of the approximation is degraded at the price of faster speed \cite{bgkl}.

Lee et al. \cite{gpis1} proposed a GP implicit surface (GPIS) map, where the distance fields are regressed instead of occupancy values. In this work, 2D GP is applied first to obtain the derivatives and 3D GP with derivative observations is followed to infer the surface distances. GPIS accurately estimates the distance fields which are close to the surface but it loses the accuracy at the regions far from the surface. Log-GPIS \cite{loggpis2} is then proposed to address this issue by applying the logarithmic transformation to a GP formulation, but it faces the  trade-offs between accuracy and interpolation abilities. Le Gentil et al. \cite{kigp} solved this problem by leveraging the relation between the kernel and the distance fields. However, the training time of those methods still scales with the increasing data size.
To reduce the time/memory complexity of GPIS, distance field priors extracted from geometry features are introduced in \cite{gpis2}. Instead of a feature extraction, this paper employed GMMs to capture the priors, which compactly parameterize point clouds, and the uncertainties are embedded in the covariance matrices of the Gaussian components.

While the GP inference, with its data-driven nature, carves local surface features, GMM favours the generalization of the whole measurement distribution. GMMs have been leveraged to generate a continuous parametric representation for point clouds\cite{hgmm, pgmm, gmm_incre, gmm_new1, gmm_new2}, infer high-fidelity occupancy\cite{gmm2, gmm3, gmm4, gmmap}, and point cloud registration\cite{hgmr, mrgmm} in the prior work. Nevertheless, model selection remains a challenge in GMM mapping for 3D point clouds\cite{pgmm}, i.e., it is difficult to determine the proper number of components for various complex scenes. Information criterion-based model selection is time-intensive as it tests various numbers to find the elbow point. Recent studies have introduced adaptive strategies that employ hierarchical structures \cite{mrgmm} and information-theoretic optimization \cite{pgmm}. However, hierarchical methods remain sensitive to hyper-parameters involved in splitting and encounter trade-offs between accuracy and efficiency \cite{mrgmm}, while SOGMM \cite{pgmm} may face the issue of selecting an optimal bandwidth parameter with the limited compute in complex outdoor scenarios.

A neural network-based method - Neural Radiance Field (NeRF) \cite{nerf} has been proposed to model the environment with a continuous implicit function of radiance and occupancy, which shows its great potential in mapping, localization, and planning. However, it lacks the sufficient well-calibrated uncertainties of the mapping results. Also, due to the inherent neural components, obtaining high-quality results continues to incur substantial expenses in training and rendering. The fixed architecture of the underlying network makes it unable to directly adapt to different scenes\cite{pgmm}. 

Recently, Gaussian splatting (GS)\cite{gs} leverages the 3D Gaussians as point-based blending with their positions and opacity, efficiently and flexibly rendering the color information compared to NeRF. Unlike GMM-mapping, GS is not explicitly calibrated to represent the point cloud distribution. Gaussians do not precisely match the actual surface \cite{sugar}. The geometry information is modelled by rendering depth images and requires additional enhancement, e.g. optical flow loss in \cite{hugs}. Also, the inherent uncertainty of Gaussians in GS is not investigated. 

In our previous work\cite{gpgmm1}, we proposed GP-mapping focusing on structured buildings, with 1D GMM as prior for the surface depth. In each local facade frame, the surface is represented in 2.5D. 
In this paper, we generalize the previous methods to more complex structures, employing 4D HGMMs to obtain the model prior and convert the local 2.5D surfaces to fully 3D representations, with fewer geometrical constraints for more general shapes. Also, the mapping process is simplified since the segmentation of local surfaces is not required anymore, reducing training time and minimizing potential segmentation errors. When GMMs are extended to higher dimensions, the issue of model selection (choice of the component number) arises. Therefore, a hierarchical GMM technique is employed to solve the selection of component numbers and speed up GMM training. GP inference is followed to correct irregular parts of GMM priors that do not comply with the original training points. 

The contributions of this work are the following:
\begin{itemize}
\item We map uncertain 3D continuous SDFs in urban areas with probabilistic inferences using laser scanning point clouds, which can be applied in various downstream tasks and can be updated probabilistically. 
The experimental results demonstrate the value of the uncertainty measures, showing that estimations with lower uncertainty lead to superior surface reconstruction outcomes.

\item  We expand the prior work\cite{gpgmm1} to a more generalized approach by applying a higher dimensional hierarchical GMM, combined with GPs to yield a fully 3D map with efficient surface prediction and to maintain a uncertainty description. This approach establishes a seamless connection between the HGMM surface distance estimation and the GP framework, contributing to enhanced accuracy, efficiency and the robustness to more general shapes. 

\item We evaluate the accuracy and the faithfulness of the uncertainty quantification with real-world LiDAR point clouds, collected by the mobile mapping system (MMS). The Root mean squared error (RMSE) and averaged predictive log-likelihood are used as metrics to evaluate the results. 
\end{itemize}

\begin{figure*}[thpb]
      \centering
      \includegraphics[width=0.79\linewidth, trim={0.67cm 0.6cm 0.65cm 0.7cm},clip]{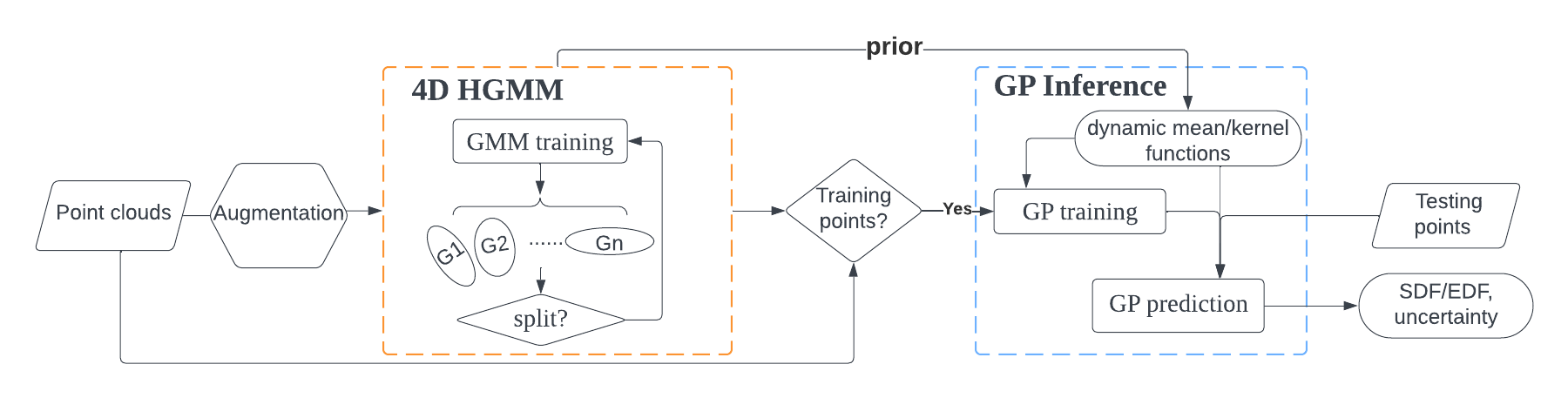}
      \caption{Overall workflow.}
      \label{fig:flowchart}
 \end{figure*}

Figure \ref{fig:flowchart} illustrates the overall mapping workflow, showing how HGMM is integrated into GP, with detailed explanations provided in the subsequent sections.

\section{Map representation}

To produce continuous maps from LiDAR observations of urban outdoor scenarios, signed-distance fields are considered as the map representation in this work, which is seen as implicit surface model, inferred by probabilistic functions. 
In a signed distance field, each point is described by a feature that represents the shortest distance to the neighboring surface. The zero-value distance denotes the boundary of object surface, while positive distances indicate the points are outside the object and negative ones are inside the object. Mathematically, this definition is described by the following functions: 
 
\begin{equation}
\label{eq:surface}
d = f(\boldsymbol{x}) \left\{ \begin{array}{cl}
= 0 & : \ \text{ on the surface}, \\
<0 & : \ \text{ inside object},  \\
>0 & : \ \text{ outside object}.
\end{array} \right.   
\end{equation}

It should be noted that in scenarios where the sign information is disregarded, one obtains the Euclidean Distance Field (EDF) as a result. EDF defines spatial relationships purely based on the magnitude of distances, absent the directional or inside-outside context provided by signed distance measures.
The surface representation can be regressed by the GMM and GP approaches and both provide the probabilistic uncertainty for the distance estimates. The GMM results serve as non-stationary prior mean and kernel functions in the GP training.

\section{Gaussian Mixture Model}
GMM-based mapping, as introduced in Section I, has been employed for occupancy probability estimation and has been proven to be a promising approach due to its flexibility from the mixture of different distributions. To solve the issue of the model selection and speed up EM optimization with large component numbers, a Hierarchical tree structure is used in this work, inspired by the previous work \cite{hgmm}. 

\subsection{Hierarchical Gaussian Mixture Models}

A standard GMM is composed of $K$ Gaussian components, described by a set of parameters $\Theta = \{(\pi_k, \boldsymbol{\mu_k},\boldsymbol{\Sigma_k})\}^K_{k=1}$, as given in:
\begin{equation}
    p(x)=\sum_{k=1}^{K}{\pi_k \mathcal{N}(x|\boldsymbol{\mu_k},\boldsymbol{\Sigma_k} )},
\end{equation}
where $\pi_k$ is the prior weight, ${\boldsymbol{\mu_k}}$ and ${\boldsymbol{\Sigma_k}}$ are the Gaussian parameters. Given a perfect $K$, the components will properly approximate the object surface. Assuming the surface can be decomposed with many planar patches, each component estimates one planar patch, and the 3-dimensional manifold degenerates into a nearly 2D structure. This implies that performing an eigenvalue decomposition on the covariance matrices of the respective components results in their smallest eigenvalues approaching zero. 

However,  a 3D GMM cannot completely degenerate into 2D. Instead, what can be achieved is the optimization of the model by targeting very small values for the principal curvatures or the eigenvalues associated with the smallest dimension. It is non-trivial to select the proper number of GMM components to achieve this goal. The idea of HGMM is to create a hierarchical tree structure for GMM models. This structure initiates modelling with a small number of Gaussian nodes to encapsulate the data. Subsequently, the complexity of each node is evaluated to select the nodes that should be split. Here, their complexity is assessed based on the deviation from a near-2D (or 'flat') structure. Nodes exhibiting a complexity beyond a predefined threshold are then subdivided into more nuanced sub-nodes at the next level of the hierarchy; i.e., new Gaussian components and GMM training are required for them. In this case, the complexity of the Gaussian node is quantified by the principal curvature, given by:

\begin{equation}
    c=\frac{\lambda_n}{\sum_{i=1}^{n}{\lambda_i}},
\end{equation}
where $\lambda_i$ is the $i$-th eigenvalue of the principal components, $\lambda_n$ is the smallest eigenvalue. If the principal curvature is small, the structure is relatively flat. The threshold is set to the same empirical value as the previous work \cite{hgmm}.

\begin{figure}[ht!]

      \begin{subfigure}{.115\textwidth}
  \centering
  \includegraphics[width=0.9\linewidth, trim={0.5cm 0cm 0.1cm 0.5cm},clip]{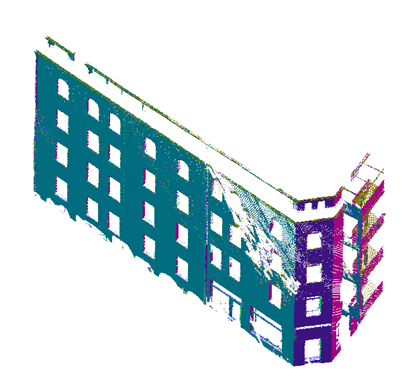}
  \caption{Data}
  \label{fig:original_data}
\end{subfigure}
\begin{subfigure}{.119\textwidth}
  \centering
  \includegraphics[width=0.9\linewidth, trim={0.7cm 0cm 0.1cm 0.5cm},clip]{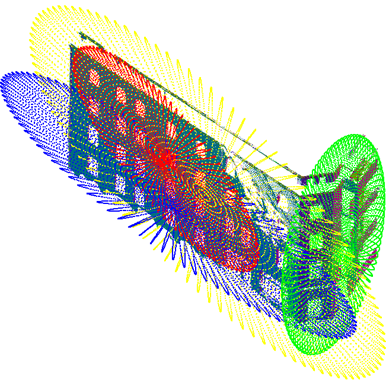}
  \caption{Level 1}
   \label{fig:level1}
\end{subfigure}
\begin{subfigure}{.119\textwidth}
  \centering
  \includegraphics[width=0.9\linewidth, trim={0.6cm 0cm 0.1cm 0.5cm},clip]{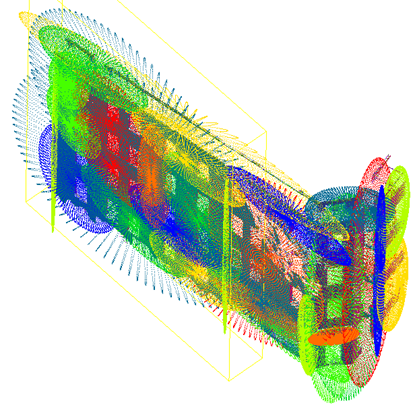}
  \caption{Level 2}
   \label{fig:level2}
\end{subfigure}
\begin{subfigure}{.121\textwidth}
  \centering
  \includegraphics[width=0.99\linewidth, trim={.6cm 0cm 0.cm 0.6cm},clip]{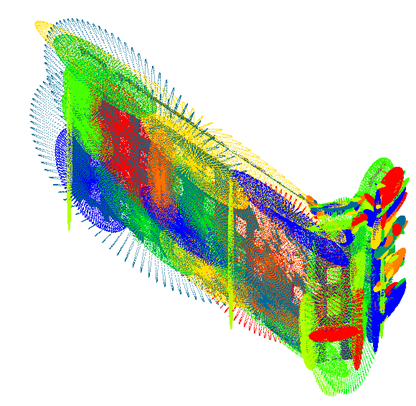}
  \caption{Level 3}
   \label{fig:finallevel}
\end{subfigure}
\caption{Visualization of the error ellipsoids of Gaussian components in HGMM. (a) is the data. (b) shows four initial Gaussians, and they split into the next level in (c). The new components are checked by the principal curvature to see if they should be further split. (d) presents the final models.}
   \label{fig:hgmm}
 \end{figure}

Figure \ref{fig:hgmm} visualizes an example where GMM structures gradually split at hierarchical levels. In the lower right regions of the building, where there are some protruded balconies, Gaussian components split from one green Gaussian (level 1) into more smaller Gaussians (level 3) to fit the local smaller surface structures. 

\subsection{Gaussian Mixture Regression}
To infer the surface distance with HGMM, the original 3D HGMM is extended to 4D space and the Gaussian mixture regression (GMR) technique is employed. 
Given LiDAR measurements ${\bf{X}}=\{{\bm{x}}_i\in \mathbb{R}^3\}_{i=1}^N$, the 3D variables (coordinates) are appended by a fourth dimension representing surface distances, denoted as $\bm{Y}=\{y_i \}_{i=1}^N$, forming the complete dataset for training: $\bm{D}=(\bf{X},\bm{Y})$. Following the EM training, the parameters $\Theta = \{(\pi_k, \boldsymbol{\mu_k},\boldsymbol{\Sigma_k})\}^K_{k=1}$ are derived and each $k$-th component is a jointly Gaussian, given by:
\begin{equation}
\boldsymbol{\mu_k}=\begin{bmatrix}
  \boldsymbol{\mu_{X}}  \\
  \boldsymbol{\mu_{Y}} 
\end{bmatrix}_k,
\boldsymbol{\Sigma_k}=\begin{bmatrix}
  \boldsymbol{\Sigma_{\bm{XX}}} & \boldsymbol{\Sigma_{\bm{X}Y}}\\
  \boldsymbol{\Sigma_{Y\bm{X}}} & \boldsymbol{\Sigma_{YY}}
\end{bmatrix}_k,
\end{equation}
\begin{equation}
\boldsymbol{\Lambda_k}=\boldsymbol{\Sigma_k}^{-1}
=\begin{bmatrix}
  \boldsymbol{\Lambda_{XX}} & \boldsymbol{\Lambda_{XY}}\\
  \boldsymbol{\Lambda_{YX}} & \boldsymbol{\Lambda_{YY}}
\end{bmatrix}_k.
\end{equation}
where $\boldsymbol{\Lambda_k}$ is the precision matrix.

In GMM regression, the rule of conditionals of multivariate Gaussian distributions is applied to estimate the queried surface distance conditioned on the 3D coordinates. For each component,  the posterior conditional distribution reads as:

\begin{equation}
    p(y|\bm{x}) = \mathcal{N}(y | \mu_{Y|\bm{X}}, \Sigma_{Y|\bm{X}} ),
\end{equation}
\begin{equation}
    \mu_{Y|\bm{X}}=\mu_{Y}-\boldsymbol{\Lambda_{YY}^{-1}}\boldsymbol{\Lambda_{Y\bm{X}}}(\bm{x}-\bm{\mu_{X}}),
\end{equation}
\begin{equation}
    \Sigma_{Y|\bm{X}} = \boldsymbol{\Lambda_{YY}^{-1}},
\end{equation}
where $\mu_{Y|\bm{X}}$ can be denoted as $ \mu_{k}(\bm{x_*})$, representing the expected surface distance for the queried point $\bm{x_*}$, and $\sigma_{k}^{2}(\bm{x_*})=\Sigma_{Y|\bm{X}}$ is the associated uncertainty.

The overall surface regression is the mixture of different Gaussian components. In fact, not all components in the GMM have effective impacts on the estimation of a queried area, rather, only some neighboring Gaussians are important, and we call these components the active components. Thus, $J$ components will be selected as active components according to the marginal probability density $p_{j}(\bm{x_*}|\bm{\mu_{jX}}, \boldsymbol{\Sigma_{\bm{jXX}}})$. 

Mixing all active Gaussian components in GMM, the regression is a mean function of all active $\mu_{j}(\bm{x_*})$ averaged with the mixing weights $w_j(\bm{x_*})$, or responsibilities, given by\cite{gmr}:
\begin{equation}
\label{eq:gmm_mean} 
m(\bm{x_*})=\sum_{j=1}^{J}{w_j(\bm{x_*}) \mu_j(\bm{x_*})},
\end{equation}
\begin{equation}
    w_j(\bm{x_*})=\frac{\pi_{j}p_{j}(\bm{x_*})}{\sum_{k=1}^{J}{\pi_{k}p_{k}(\bm{x_*})}},
\end{equation}
and the uncertainty of the expected surface distances is specified as the variance or standard deviation:
\begin{equation}
\label{eq:gmm_sigma}
\sigma^{2}(\bm{x_*})={\sum_{j=1}^{J}{w_{j}(\bm{x_*})\{\sigma_j^{2}+\mu_j(\bm{x_*})^{2}\}-m(\bm{x_*})^{2}}}.
\end{equation}

As illustrated above, the signed distance value $y$ of each data point is required in the training process. However, the measurements are all collected when the scanning ray hits the surface, i.e., $y=0$ for all measurements. To obtain valid covariances between spatial coordinates and surface distances, the original dataset is augmented with virtual points along the normal direction of the hit points and at a small distance from the surface. The extension of GMM dimension for map regression was originally proposed in the prior work \cite{gmm2}. The key differences are: 1) the proposed method applies a “top-down” strategy with faster computation speed; 2) SDF is regressed instead of occupancy. 

Despite the extension of the hierarchical model from 3D to 4D and the consequent changes in the interpretations of complexity and split criteria, these criteria remain applicable and suitable in the 4D context. When the smallest eigenvalue approaches zero, it indicates that the variance along one principal axis is minimal. The N-dimensional manifolds are nearly degenerated into (N-1)-dimensional manifolds. This condition typically leads to a tighter correlation between the variables involved. Consequently, in our case, the regression of the fourth dimension (surface distances) based on point coordinates will likely exhibit less noise and can be expected to be more accurate. In addition, it often forces the first 3D variables to have a more flat structure as well. 

Although the hierarchical strategy helps us to adaptively find  the number of Gaussian components  and quickly optimize them, the performance is still sensitive to the thresholds of the complexity and the size of the modeling objects. Using a too-big threshold sometimes ignores the details of large objects, while it might oversplit the nodes with a too-small threshold. Also, one single threshold may affect the ability to fit the scene with both large and small objects and also the threshold can become too small with a higher hierarchy of smaller nodes. Therefore, HGMM is an approximation of the ideal GMM with the best fidelity, which can be further improved. Rather than selecting the perfect number of components, GP-based mapping is followed to refine the results from HGMM. More specifically, parameters optimized in HGMM are used as priors for GP training, which will be introduced in Section~IV.

\section{Gaussian Processes with HGMM priors}
\subsection{Gaussian Processes} 
The GP model $f \sim \mathcal{GP} (\mu(\boldsymbol{x}),k(\boldsymbol{x},\boldsymbol{x}'))$, as a flexible non-parametric approach, can be seen as a distribution of functions. It consists of two functions: a mean function and a covariance function (kernel). Given the input data $\boldsymbol{x}$, the mean function defines the expectation of the target value $y=\mu(\boldsymbol{x})$, while the kernel function determines the variance $k(\boldsymbol{x},\boldsymbol{x})$ of this target value and its covariance $k(\boldsymbol{x},\boldsymbol{x}')$ with another $y'$ at any other input location $\boldsymbol{x}$.

Supposing a training dataset with location inputs ${\bf{X}}=\{{\bm{x}}_i \in \mathbb{R}^3\}_{i=1}^N$ and the corresponding target distance values $\bm{y}=\{y_i \}_{i=1}^N$ measured with additive noise $\eta_i \sim \mathcal{N}(0,\sigma_\eta^2)$. The inference of the testing space $\bf{X}_*$ can be estimated from the posterior GP mean and kernel \cite{gp-book}:

\begin{equation}
\label{eq:gpm}
\mu(\boldsymbol{x}_*)=\mu_0(\boldsymbol{x}_*)+ {\bf k}_*^T ({\bf K}_N+\sigma_\eta^2 {\bf I} )^{-1} (\boldsymbol{ y}-\mu_0 ({\bf X})),
\end{equation}
\begin{equation}
\label{eq:gpv}
k(\boldsymbol{x}_*,\boldsymbol{x}_*')=k_0 (\boldsymbol{x}_*,\boldsymbol{x}_*') - {\bf k}_*^T ({\bf K}_N+\sigma_\eta^2 {\bf I} )^{-1} {\bf k}_*'+\sigma_\eta^2,	
\end{equation}
where $\mu_0 (\boldsymbol{x})$ and $k_0 (\boldsymbol{x},\boldsymbol{x}')$ are denoted as the prior mean and kernel function. ${\bf K}_N= k_0 ({\bf{X}}, {\bf{X}})$ is the covariance matrix of the $N$ input points calculated by the prior kernel function. $[{\bf k}_*]_N= k_0 ({\bf X}, \boldsymbol{x}_* )$ and  $[{\bf k}_*' ]_N= k_0 ({\bf X}, \boldsymbol{x}_*' )$ are the prior covariance between the $N$ input training points and the predicting points.

In a standard stationary GP, the prior mean function is often set to a constant number, e.g., zero, and the kernel function depends only on the distance between points in the input space and not on the specific locations of those points, e.g., the squared exponential kernel or Matérn kernel. For instance, a $3/2$ Matérn kernel used as the prior GP kernel function reads as:

\begin{equation}
\label{eq:kernel} 
k_0(\boldsymbol{x},\boldsymbol{x}')=\sigma_p^2 \left(1 + \frac{\sqrt{3}\left | \boldsymbol{x}-\boldsymbol{x}' \right | }{\ell}\right)\exp\left(-\frac{\sqrt{3}\left | \boldsymbol{x}-\boldsymbol{x}' \right | }{\ell}\right), 
\end{equation}
where
$\sigma_p$ is the prior standard deviation and $l$ is the length scale parameter. 

As a stationary kernel, the hyper-parameters $\sigma_p$ and $l$ in the above equation should remain invariant at different input locations. In this paper, however, the prior mean and prior variance come from the outcomes of HGMM, which vary across different inputs. It indicates that the stationary GP assumption does not hold anymore and the GP regression transforms into a non-stationary case.

\subsection{GP Regression with Non-stationary Priors from HGMM} 

To integrate both methodologies seamlessly, we utilize surface distance estimation derived from HGMM as the prior mean for GP. The GMM regression function (Equation \eqref{eq:gmm_mean}) is employed as the prior mean function during the training of the GP: $\mu_0(\boldsymbol{x}) = m(\bm{x})$.

\begin{figure}[ht!]
\begin{subfigure}{.2215\textwidth}
  \centering
  \includegraphics[width=0.8\linewidth]{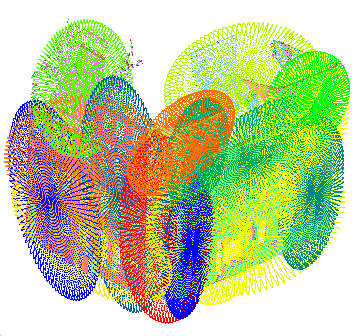}
  \caption{GMM components}
  \label{fig:gmm_comps}
\end{subfigure}
\begin{subfigure}{.265\textwidth}
\centering
  \includegraphics[width=0.8\linewidth]{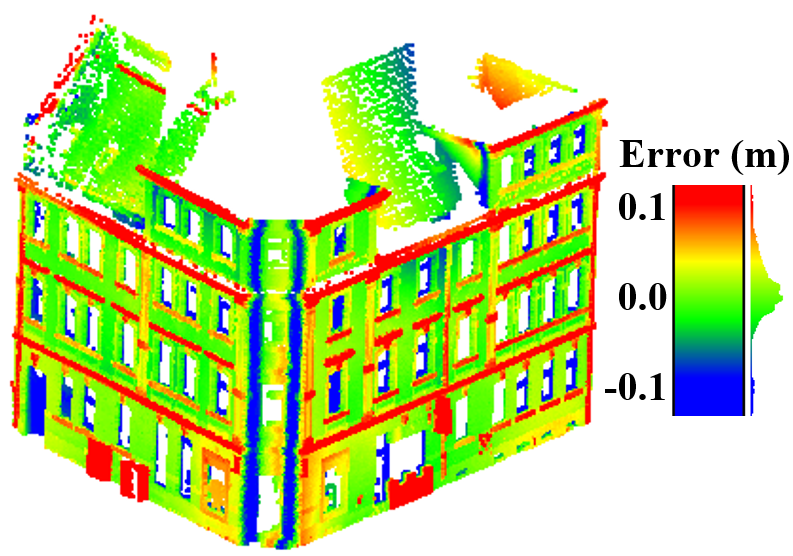}
  \caption{GMM surface}
  \label{fig:gmmsurf}
\end{subfigure}
\begin{subfigure}{.21\textwidth}
  \centering
  \includegraphics[width=0.8\linewidth]{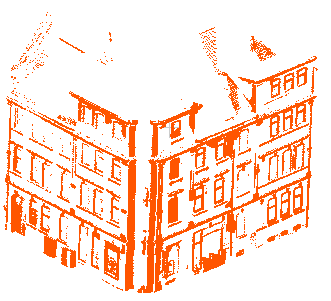}
  \caption{Training points}
   \label{fig:train_pts}
\end{subfigure}
\begin{subfigure}{.265\textwidth}
  \centering
  \includegraphics[width=0.8\linewidth]{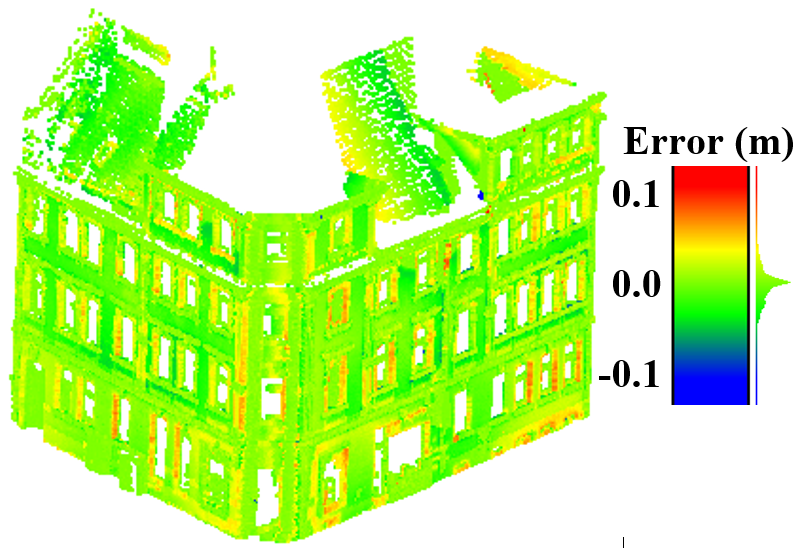}
  \caption{GP surface}
   \label{fig:gpsurf}
\end{subfigure}

\caption{Selected training points for GP: (a) shows 16 Gaussian components of GMM. (b) presents the differences between estimated signed distances and measured surface points. The green color indicates small errors as shown in the color bar. The points with large errors are selected for GP training in (c). (d) shows the GP inference results.}
   \label{fig:GMMerror}
 \end{figure}
 
Accordingly, the estimated standard deviation computed in Equation \eqref{eq:gmm_sigma} in Section III serves as $\sigma_p$ in the kernel function. Since the hyper-parameter $\sigma_p$ varies with different inputs, Equation \eqref{eq:kernel} also changes into:

\begin{equation}
\label{eq:kernel2} 
k_0(\boldsymbol{x},\boldsymbol{x}')=\sigma_p\sigma_p^{'}  \left(1 + \frac{\sqrt{3}\left | \boldsymbol{x}-\boldsymbol{x}' \right | }{\ell}\right)\exp\left(-\frac{\sqrt{3}\left | \boldsymbol{x}-\boldsymbol{x}' \right | }{\ell}\right), 
\end{equation}

To fully exploit the HGMM results obtained from HGMMs and reduce the model size of GP training, we select only the points that are not well-modelled on the surface to be the training points, as shown in Figure \ref{fig:GMMerror}. The selection criteria is given by:
\begin{equation}
\label{eq:condition1}
|m(\bm{x}) |>d_m,
\end{equation}
where $d_m$ is the allowed maximal discrepancy between the real observation and the HGMM estimation of surface distance. As measurements hit the surface, the real observation is always zero. This criterion indicates the HGMM estimation should be smaller than a given threshold. Otherwise, the point will be selected for GP training. This value is dependent on the required accuracy for the surface. 

\subsection{GP with Derivative Observations} 

Leveraging derivatives of the targets (distance fields) presents a potentially valuable approach for enforcing known constraints \cite{gp-book}; i.e., a more robust inference of the signed distance fields can be made based on the gradients of the distances.  In this case, the normal vectors of the points represent the derivative observations at the local surface. The distances increase along the normal direction.

Therefore, we augment the original 4D training dataset with additional normal information about each point, yielding a new dataset $\mathcal{D}=\{\bm{x_i}, y_i, \nabla \bm{y_i}\}_{i=1}^N$. The inference can be made based on the joint Gaussian distribution of the signed distances and the gradients of distances. Accordingly, the joint covariance matrix $\tilde{k}(\bm{x}, \bm{x}')$ involves $N(3+1)$ entries\cite{loggpis2, gp-book}:

\begin{equation}
\label{eq:kernel_j}
\tilde{k}(\bm{x}, \bm{x}') = 
\begin{bmatrix}
  k(\bm{x}, \bm{x}') & k(\bm{x}, \bm{x}')\nabla_{\mathbf{x'}} ^\top  \\
  \nabla_{\mathbf{x}} k(\bm{x}, \bm{x}') &\nabla_{\mathbf{x}} k(\bm{x}, \bm{x}')\nabla_{\mathbf{x'}} ^\top
\end{bmatrix},
\end{equation}
where $\nabla_{\mathbf{x}} k(\bm{x}, \bm{x}')$ is the gradient of the covariance function with respect to $\bm{x}$, i.e., the covariance between target values and the partial derivatives. $\nabla_{\mathbf{x}} k(\bm{x}, \bm{x}')\nabla_{\mathbf{x'}} ^\top$ is the covariance between partial derivatives. 

The gradients or the normal vectors are estimated by PCA for each point. Given the training data $\mathcal{D}$ with derivative information, we can change the standard inference, substituting $k_0(\cdot, \cdot)$ in Equation \eqref{eq:gpm} and \eqref{eq:gpv} by Equation \eqref{eq:kernel_j}:

\begin{equation}
\label{eq:gpm2}
\mu(\boldsymbol{x}_*) = 
\begin{bmatrix} 
m(\boldsymbol{x}_*)\\
\nabla m(\boldsymbol{x}_*)
\end{bmatrix} + {\bf \tilde{k}}_*^T ({\bf \tilde{K}}_N+\sigma_\eta^2 {\bf I} )^{-1} 
\begin{bmatrix}
\boldsymbol{ y}-m ({\bf X})\\
\nabla \bm{y} - \nabla m(\bf{X})
\end{bmatrix},
\end{equation}
\begin{equation}
\label{eq:gpv2}
k(\boldsymbol{x}_*,\boldsymbol{x}_*')=\tilde{k} (\boldsymbol{x}_*,\boldsymbol{x}_*') - {\bf \tilde{k}}_*^T ({\bf \tilde{K}}_N+\sigma_\eta^2 {\bf I} )^{-1} {\bf \tilde{k}}_*'+\sigma_\eta^2,	
\end{equation}

The technique of GP inference with derivative observations has been employed in the existing work \cite{gpis1, loggpis2}. They applied additional 2D observation GP and occupancy tests to estimate the normals of measurement points. Unlike that, normal vectors are estimated by PCA directly here.

\section{Experiments and Evaluation}
In this section, the datasets and experimental results are illustrated. The SDF and EDF estimated from the proposed approach are compared with the existing state-of-the-art methods: GPIS \cite{gpis1} and Log-GPIS \cite{loggpis2}. We also compare the differences between using GMM alone and GP-inference with GMM priors (GMMGP). The work is implemented in C++ and Python on a CPU of Intel i9-11900 @ 2.50GHz.
 
\subsection{Dataset}

The experimental evaluation of the proposed method utilizes real 3D LiDAR point clouds representing intricate urban environments, sourced from the LUCOOP dataset \cite{lucoop}. These point clouds were acquired by the Riegl VMX-250 mobile mapping system, operating within Hannover, Germany. To ensure the generation of densely mapped point clouds suitable for both modelling and testing purposes, the equipped van traversed the designated route on multiple measurement campaigns. Thus, the point clouds have a sufficient density (mean nearest distance = 5mm) to facilitate the generation of ground truth for the SDF. This involves computing distances between a query point and its nearest neighbor within the dense reference point clouds. Since the reference points are not continuous surfaces but rather discrete samples, it is essential to make them densely packed to minimize errors in the ground truth. Approximately 10\% of the entire dataset is selected as training data.

\begin{figure}[ht!]
  \centering
  \includegraphics[width=0.7\linewidth]{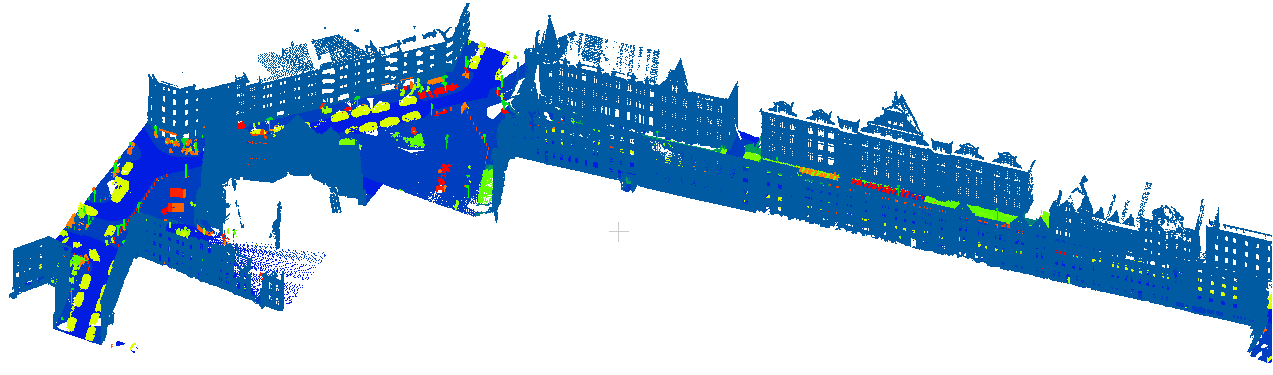}
  \caption{Urban point clouds in the LUCOOP dataset \cite{lucoop}: colors are the labels of different Geo-objects.}
   \label{fig:data}
 \end{figure}

\subsection{Practical Implementation}

In the practical implementation of HGMMs, the original observations are augmented with virtual points from both sides of the surfaces. To speed up training, we sub-sample the point clouds and virtual points are generated from these sub-sampled points, which is around $15\%$ of the original size. We trained the model with the subsampling rates of $5\%, 15\%, 40\%, 80\%$, and the results show no significant differences. The HGMM approach is robust for the sparse data. 

The spacing between virtual points and hit points is set to 0.25m, based on multiple considerations. The confidence in the virtual distance decreases with increasing separation from the hit point, suggesting that the designated space should not be too large. However, the small spacing restricts the model's ability to capture distance field gradients at greater distances. Empirical results indicate that a reduced interval enhances the accuracy near the surface while leading to worse estimations at further distances. The determination of the spacing is a trade-off of the above considerations.

Also, the GP inference with derivatives faces scalability challenges with large datasets due to its computational complexity, which scales cubically $\mathcal{O}(N^3D^3)$ with the size of the data. To address this, a pragmatic implementation involves partitioning the data into blocks organized in an Octree \cite{octomap} structure, the same as applied in the GPIS map \cite{gpis1}. 
 
\subsection{Surface reconstruction}
 
\begin{figure}[ht!]
      
\begin{subfigure}{.15\textwidth}
  \centering
  \includegraphics[width=0.88\linewidth]{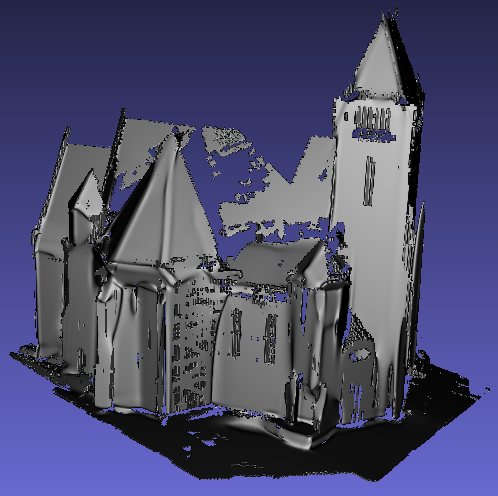}
  \caption{}
  \label{fig:qualitative_gmm}
\end{subfigure}
\begin{subfigure}{.15\textwidth}
  \centering
  \includegraphics[width=0.9\linewidth, trim={0cm 0cm 0.cm 0cm},clip]{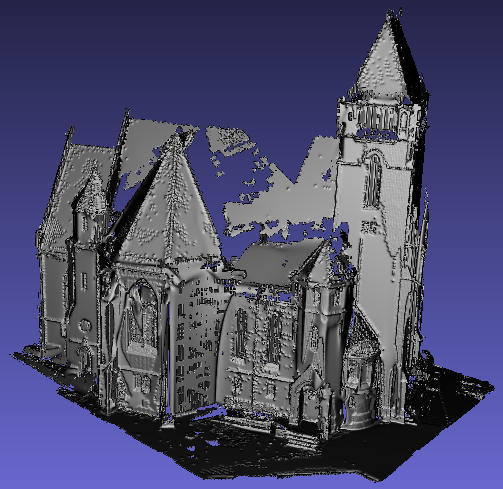}
  \caption{}
   \label{fig:qualitative_gp}
\end{subfigure}
\begin{subfigure}{.15\textwidth}
  \centering
  \includegraphics[width=0.92\linewidth, trim={0cm 0cm 0.cm 0.cm},clip]{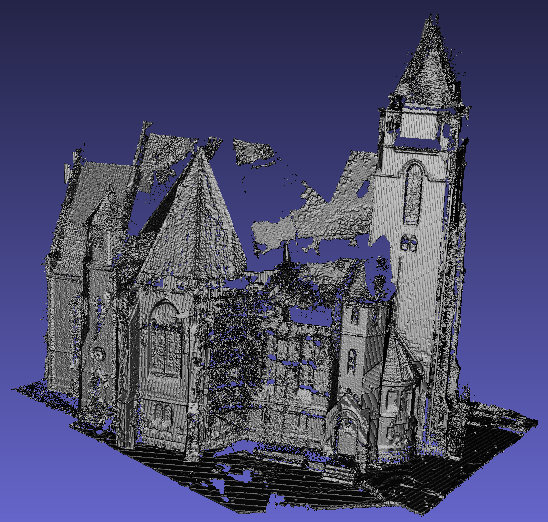}
  \caption{}
  \label{fig:qualitative_gpgmm1}
\end{subfigure}

\begin{subfigure}{.15\textwidth}
  \centering
  \includegraphics[width=0.88\linewidth, trim={0.18cm 0.2cm 0.2cm 0.15cm},clip]{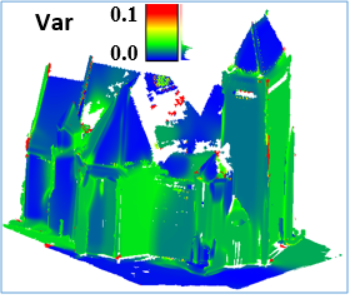}
  \caption{}
  \label{fig:qualitative_gmmv}
\end{subfigure}
\begin{subfigure}{.15\textwidth}
  \centering
  \includegraphics[width=0.9\linewidth, trim={0.1cm 0.2cm 0.3cm 0.15cm},clip]{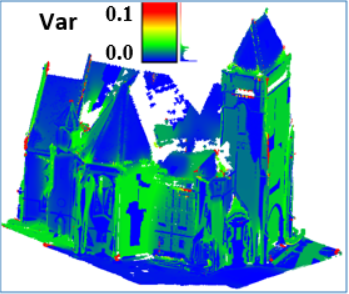}
  \caption{}
   \label{fig:qualitative_gpv}
\end{subfigure}
\begin{subfigure}{.15\textwidth}
  \centering
  \includegraphics[width=0.91\linewidth, trim={0.1cm 0.2cm 0.2cm 0.2cm},clip]{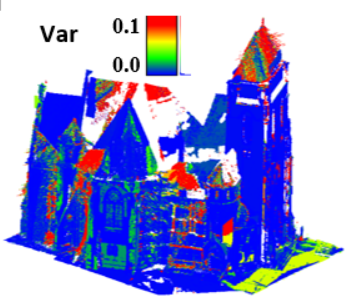}
  \caption{}
   \label{fig:qualitative_gpgmmv}
\end{subfigure}

\caption{Qualitative results extracted by the marching cubes algorithm: (a) surface of 
 HGMM; (b) surface of GMMGP; (c) surface of the previous GMMGP2.5D. (d), (e) and (f) are the associated uncertainty (variance) of the HGMM/GMMGP/GMMGP2.5D surfaces; As shown by the color bars, redder colors indicate larger uncertainties and blue colors denote small uncertainties.}
   \label{fig:qualitative}
 \end{figure}

In Figure \ref{fig:qualitative}, the qualitative outcomes of the mapping are presented with their corresponding uncertainties. For visualization purposes, the implicit surface undergoes post-processing via the marching cubes algorithm\cite{mc} to render the explicit surfaces at the zero crossing of the signed distance fields. Notably, the surfaces derived from the integrated GMMGP approach (Figure \ref{fig:qualitative_gp}) exhibit enhanced detail compared to those generated using HGMM alone (Figure \ref{fig:qualitative_gmm}). Furthermore, regions exhibiting a deficiency in detail correspondingly display a higher uncertainty on the HGMM surface. Following the GP correction, the uncertainty measures in those areas decreased. Therefore, the GMMGP results show broader areas with lower uncertainty, indicating an improvement in the overall reliability of the model. The prior work (GMMGP2.5D) \cite{gpgmm1} uses segmented local 2.5D frames, resulting in increased surface discontinuity in reconstruction. It performs better in primarily planar areas but occasionally presents discontinuous reconstruction errors. Moreover, it generally underestimates the uncertainty relative to HGMM and GMMGP.

\subsection{Evaluation}
To evaluate both the mapping accuracy and the reliability of the uncertainty measures, the proposed GMMGP method is benchmarked with existing state-of-the-art mapping approaches: GMMGP2.5D \cite{gpgmm1}, GPIS map \cite{gpis1}, Log-GPIS \cite{loggpis2} and Kernel-Inverse GP (KIGP) \cite{kigp}, which all provide uncertainty measures along with the mapping results. The uncertainty measure of KIGP is a a proxy and does not align with our quantitative evaluation, hence it is excluded from comparison. We compared the SDF of the testing points between GMMGP2.5D, GPIS and the proposed HGMM and GMMGP. Given that Log-GPIS and KIGP only provide unsigned distance fields, EDF was compared. 

There are two metrics used for evaluation: 1)~Root Mean Squared Error (RMSE) for the accuracy of the distance field; 2)~Predictive log-likelihood for the uncertainty evaluation. RMSE is computed between the ground truth and the resultant distances from the methods across a set of testing points. Predictive log-likelihood is a popular metric to evaluate uncertainty quality, quantifying how well the estimated distribution accords with the true distribution. The larger measure indicates a better uncertainty quality. In this work, log-likelihood values are averaged by the number of testing samples, given by:

\begin{equation}
    LL = \frac{1}{N_*} \sum_{i=1}^{N_*} \log p(\mu_i|\hat{\mu_i}, \hat{\sigma}^{2}_i)
\end{equation}
where $N_*$ is the number of the testing points $\bf{X_*}$, $\mu$ is the ground truth distance, $\hat{\mu}$ and $\hat{\sigma}^2$ are the estimation and posterior uncertainty obtained from Equation \eqref{eq:gpm2} and \eqref{eq:gpv2} .

\begin{figure}[ht!]
\begin{subfigure}{.241\textwidth}
  \centering
  \includegraphics[width=.98\linewidth, trim={0.cm 0.cm 0.8cm 0.8cm},clip]{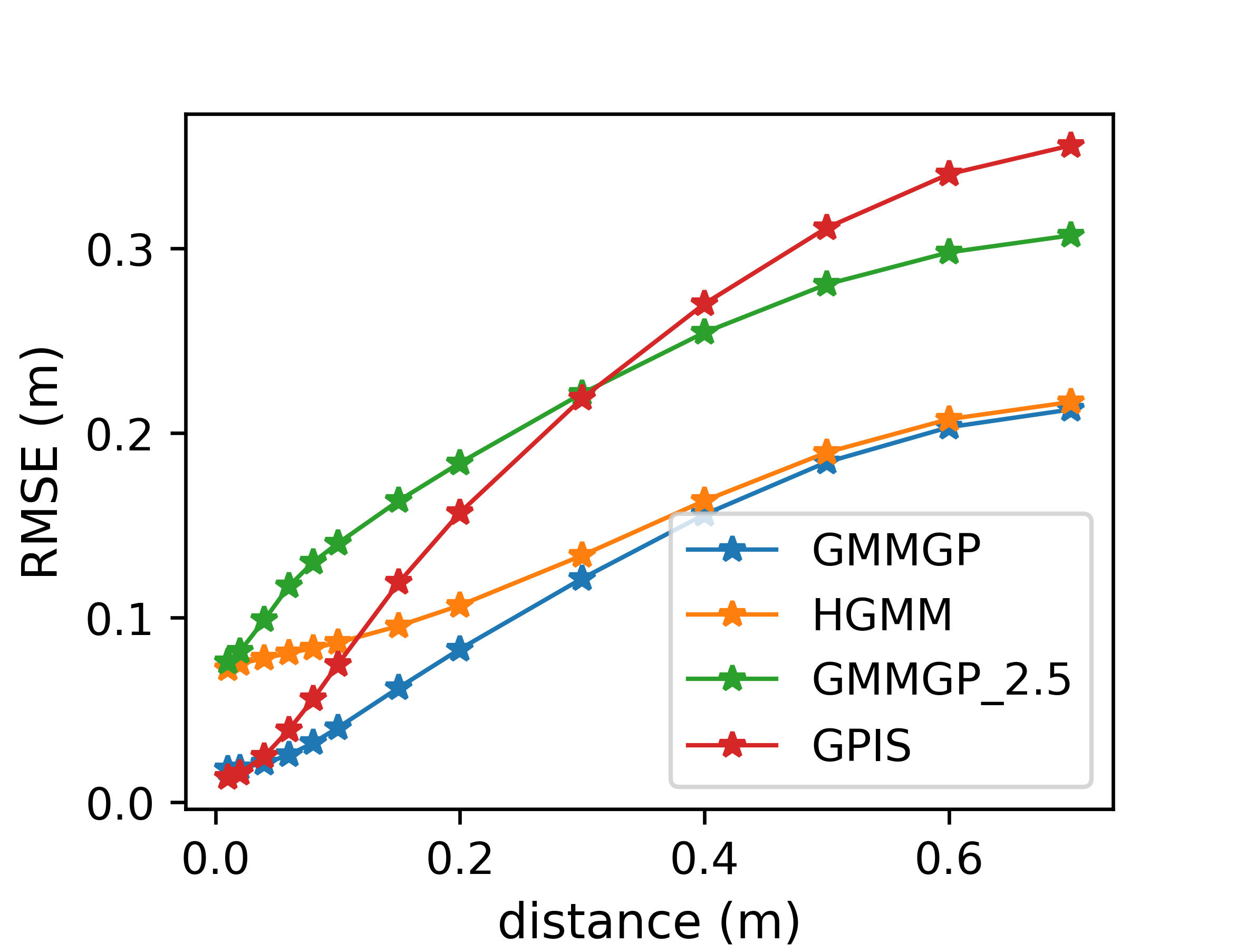}
  \caption{SDF}
  \label{fig:rsdf}
\end{subfigure}
\begin{subfigure}{.241\textwidth}
  \centering
  \includegraphics[width=.98\linewidth, trim={0.05cm 0.cm 0.9cm .8cm},clip]{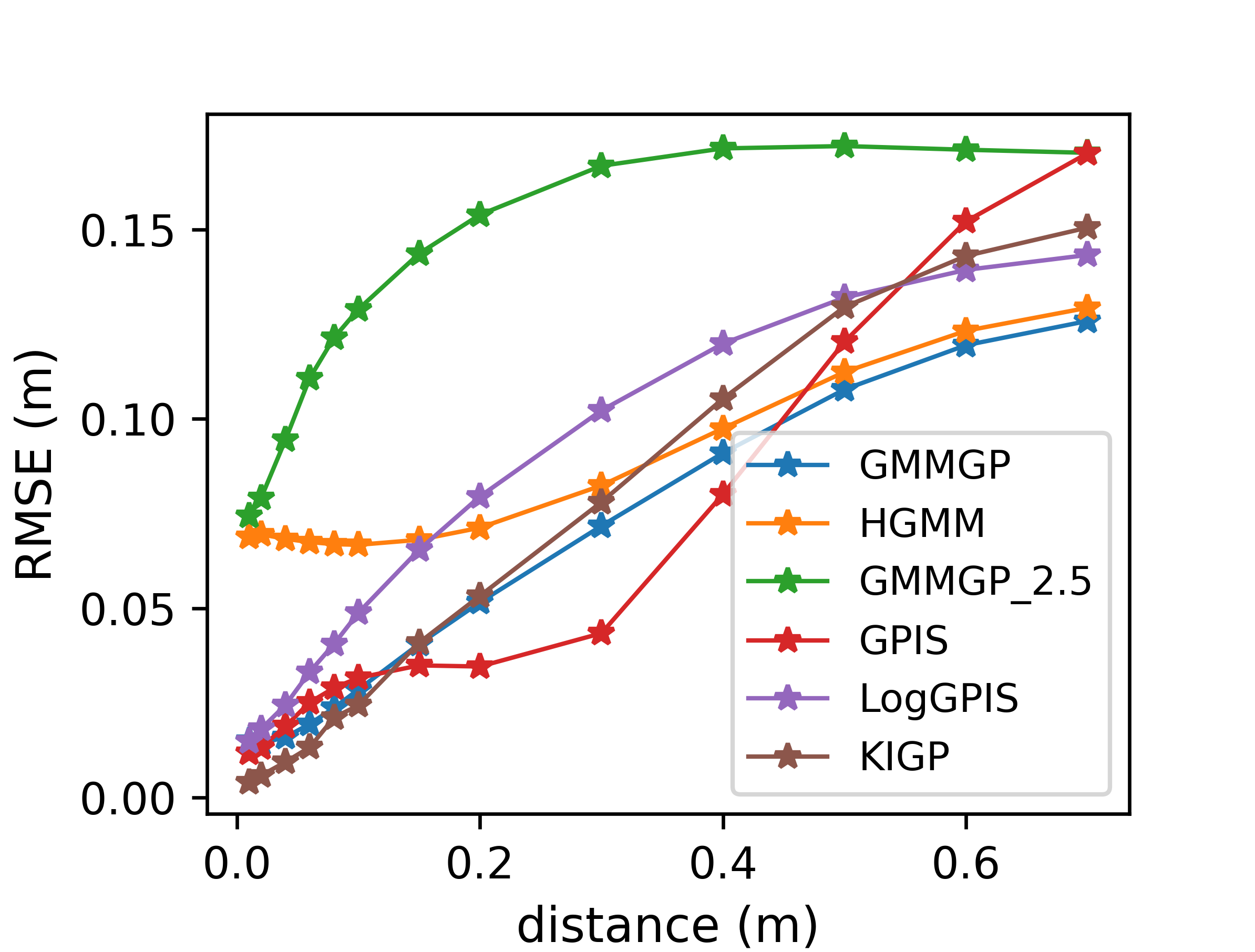}  
  \caption{EDF}
  \label{fig:redf}
\end{subfigure}

\caption{Accuracy evaluation with RMSEs.}
\label{fig:rmse}
\end{figure}

\begin{figure}[ht!]
\begin{subfigure}{.241\textwidth}
  \centering
  \includegraphics[width=.96\linewidth, trim={0.11cm 0.cm 0.9cm 0.8cm},clip]{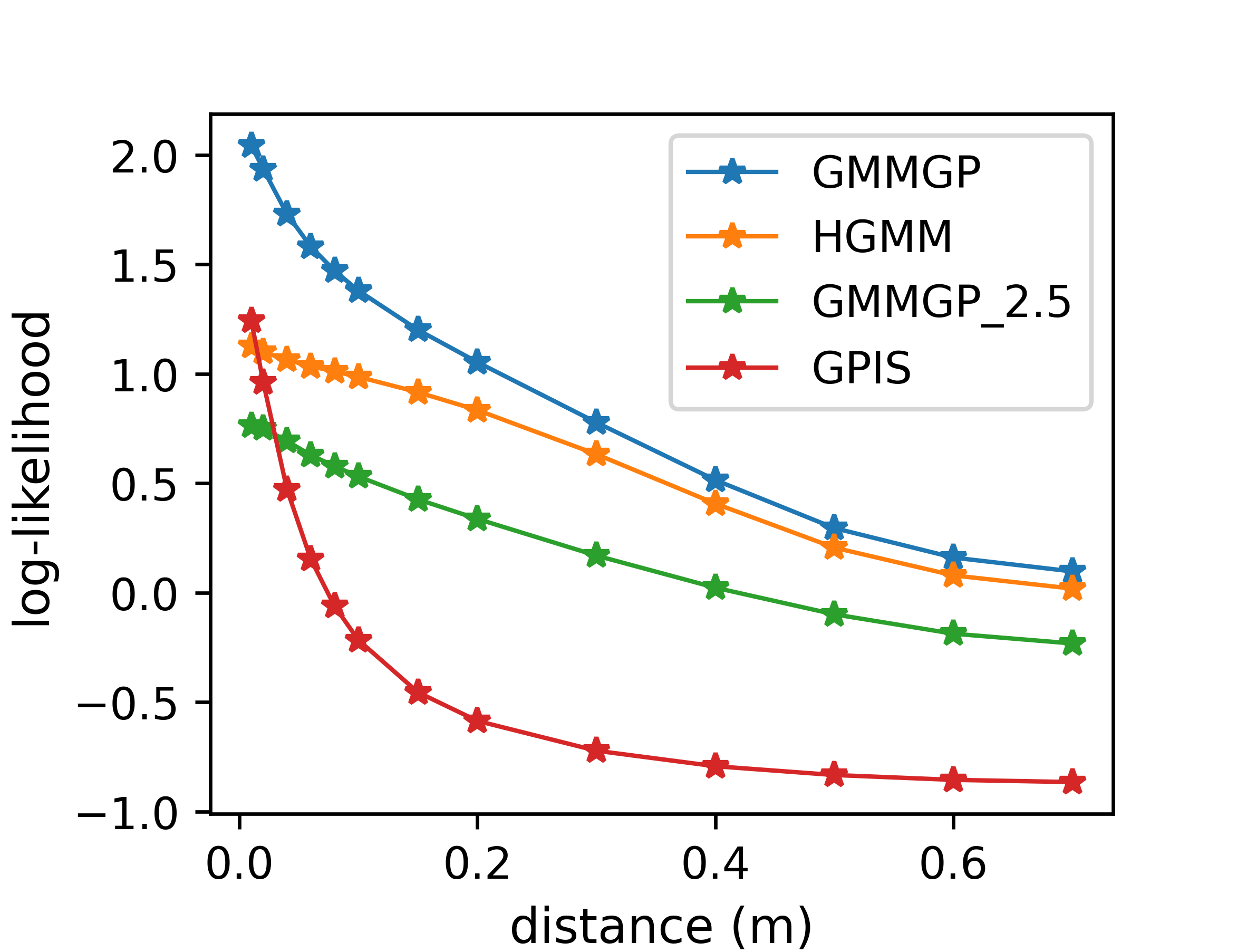}
  \caption{SDF}
  \label{fig:lsdf}
\end{subfigure}
\begin{subfigure}{.241\textwidth}
  \centering
  \includegraphics[width=.97\linewidth, trim={0.cm 0.cm 0.9cm 0.8cm},clip]{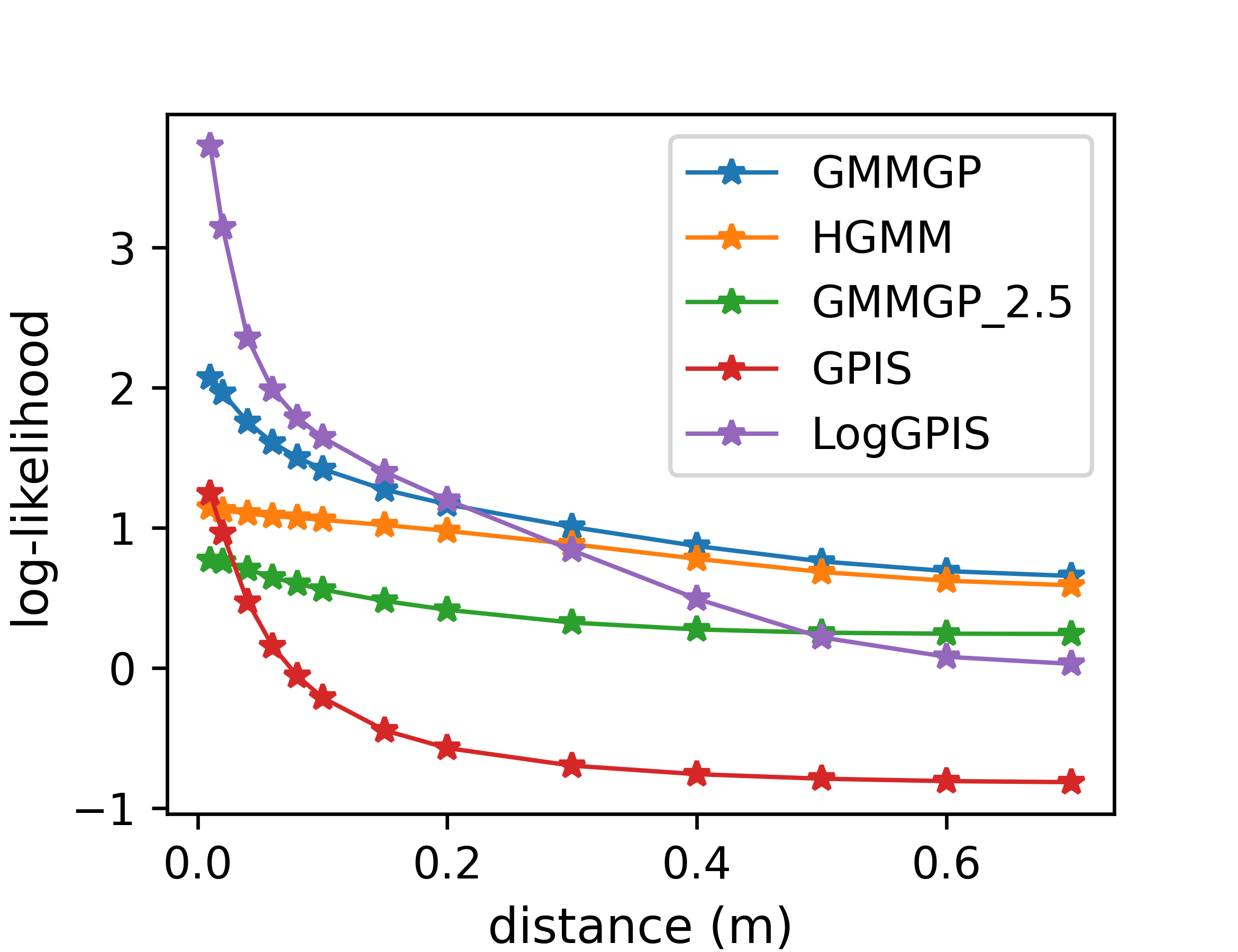}
  \caption{EDF}
  \label{fig:ledf}
\end{subfigure}

\caption{Uncertainty evaluation with Log-likelihood.}
\label{fig:ll}
\end{figure}

Figure \ref{fig:rmse} and \ref{fig:ll} present a comprehensive comparison of RMSEs and log-likelihoods for estimated surface distances, encompassing both established benchmark methodologies and our proposed GMMGP as well as standalone HGMM mapping. Both metrics are plotted against varying distances between the query points and the original training points. With a growing distance from the observations, a decline in performance is noted across all methods, characterized by increasing RMSEs and diminishing log-likelihoods. 


In the analysis of the SDF accuracy (Figure \ref{fig:rsdf}), GMMGP exhibits the best performance with the smallest RMSEs across the entire range of distances. The discrepancy between GMMGP and HGMM is particularly pronounced in regions proximal to the surface. Here, GMMGP demonstrates an enhanced ability to accurately infer the surface and its adjacent distance fields - a distinction in the qualitative results is compared in Figure \ref{fig:qualitative}, where Figure \ref{fig:qualitative_gp} shows more details on the reconstructed surface than \ref{fig:qualitative_gmm}. Since the GMMGP2.5D is proposed for structured building models, more arbitrary and complicated shapes can cause improper segmentation for local planes. It shows worse robustness and accuracy than the proposed GMMGP in full 3D. Although GPIS shows accurate inference for query points near the surface, its RMSEs notably increase as the distance from the surface increases. Mirroring this pattern, GMMGP tends to align with the GMM's outcomes for query points located at larger distances, indicating a convergence towards the prior model in areas far from the observational data.

Within the context of EDF outcomes, GMMGP generally achieves higher accuracy. For distances less than 0.1m, the outcomes from the KIGP are marginally superior. At proximities close to 0.2m, GPIS shows the lowest RMSEs. This result arises because GPIS tends to converge towards the prior beyond a certain threshold, with the experimental setup defining this prior distance as 0.2 meters. Given that signs are disregarded in EDF calculations, this prior setting intrinsically aids GPIS in obtaining accurate estimations for query points around the 0.2-meter mark. However, as distance increases, the error associated with GPIS exceeds that of the proposed GMMGP method, highlighting the limitations of relying on the fixed prior in distance estimation tasks.

In the assessment of uncertainty, the proposed GMMGP demonstrates promising performance. When evaluating SDF, both the GMMGP and HGMM outperform GPIS and the previous work GMMGP2.5D, exhibiting higher likelihoods. For EDF, the uncertainty quantification provided by Log-GPIS proves to be more dependable at proximal distances, with log-likelihoods diminishing rapidly as distance increases. Overall, the GMMGP method exhibits the most consistent and robust performance relative to the alternative approaches considered.

The predetermined priors in GPIS lead to diminished accuracy at greater distances from observed data. To solve this problem, log-GPIS introduced a heat model in the standard GPIS, which modifies the regression target as the exponential transformation of the original surface distances. This converts the target value at the surface from zero into one, and transitions the prior mean from a fixed value to infinity $\sim -\log (0)$. However, Log-GPIS encounters constraints related to the choice of the length scale parameter; e.g., a small length scale parameter leads to more accurate results at the expense of surface smoothness. Additionally, the small length scale parameter also limited its capability to infer further distance.
Although KIGP mapping employs kernel reverting functions to improve Log-GPIS, it is not scalable to the large 3D dataset as it applies full GP inference. Thus, in this case, the data must be partitioned into  cells for large urban scenes, which, however, affects its accuracy.

In contrast, our proposed method addresses these challenges by integrating the HGMM to obtain a calibrated prior, and employing local GPISs with non-stationary mean and kernels. This enhances the accuracy of signed distance field estimations and achieves more dependable uncertainty measures. 

\subsection{Computation time comparison}

\begin{figure}[ht!]
\begin{subfigure}{.241\textwidth}
  \centering
  \includegraphics[width=.95\linewidth, trim={0.0cm 0.cm 0.95cm 0.85cm},clip]{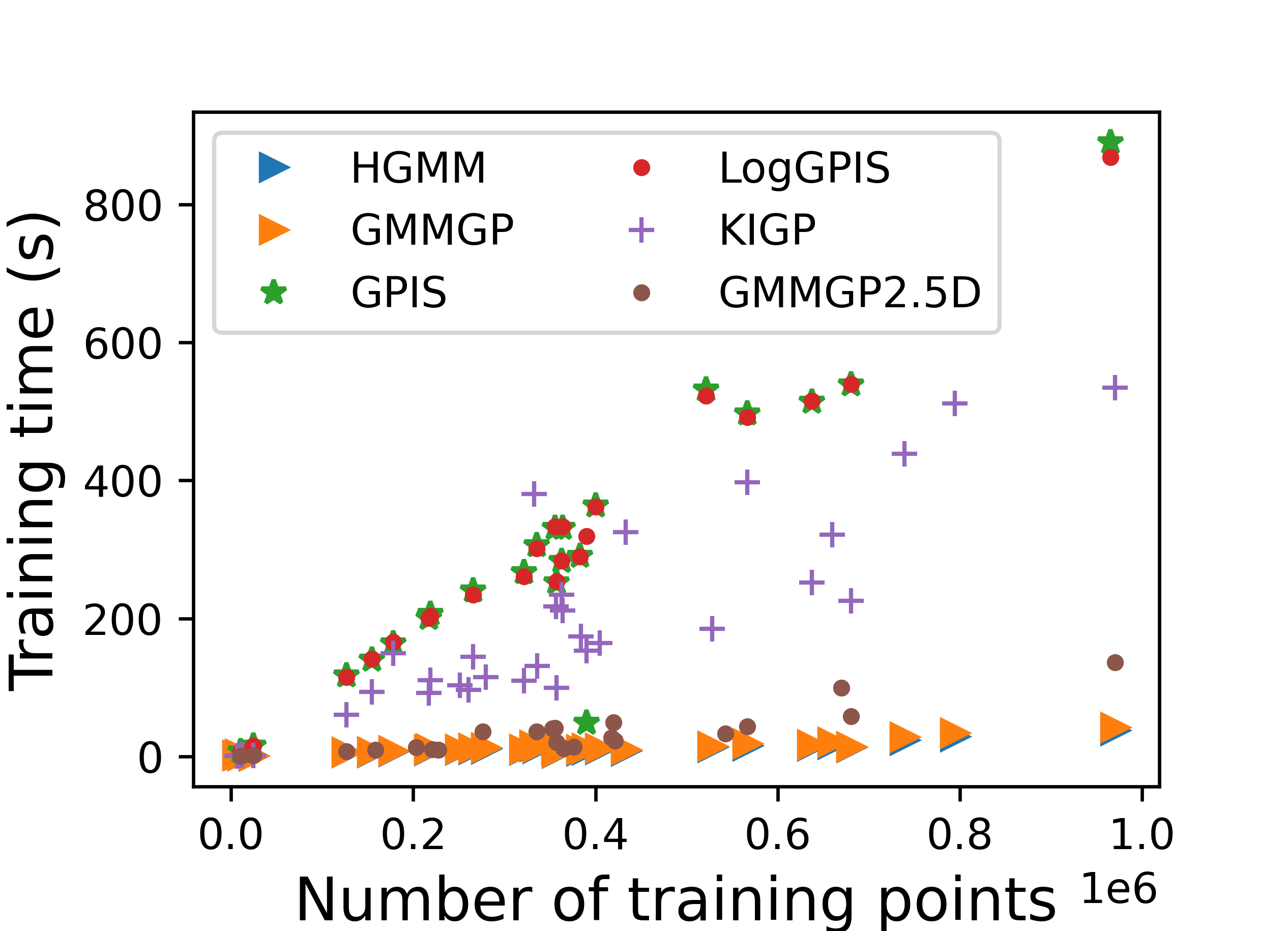}  
  \caption{Training time}
  \label{fig:train1} 
\end{subfigure}
\begin{subfigure}{.241\textwidth}
  \centering
  \includegraphics[width=.95\linewidth, trim={0.cm 0.cm 0.95cm 0.85cm},clip]{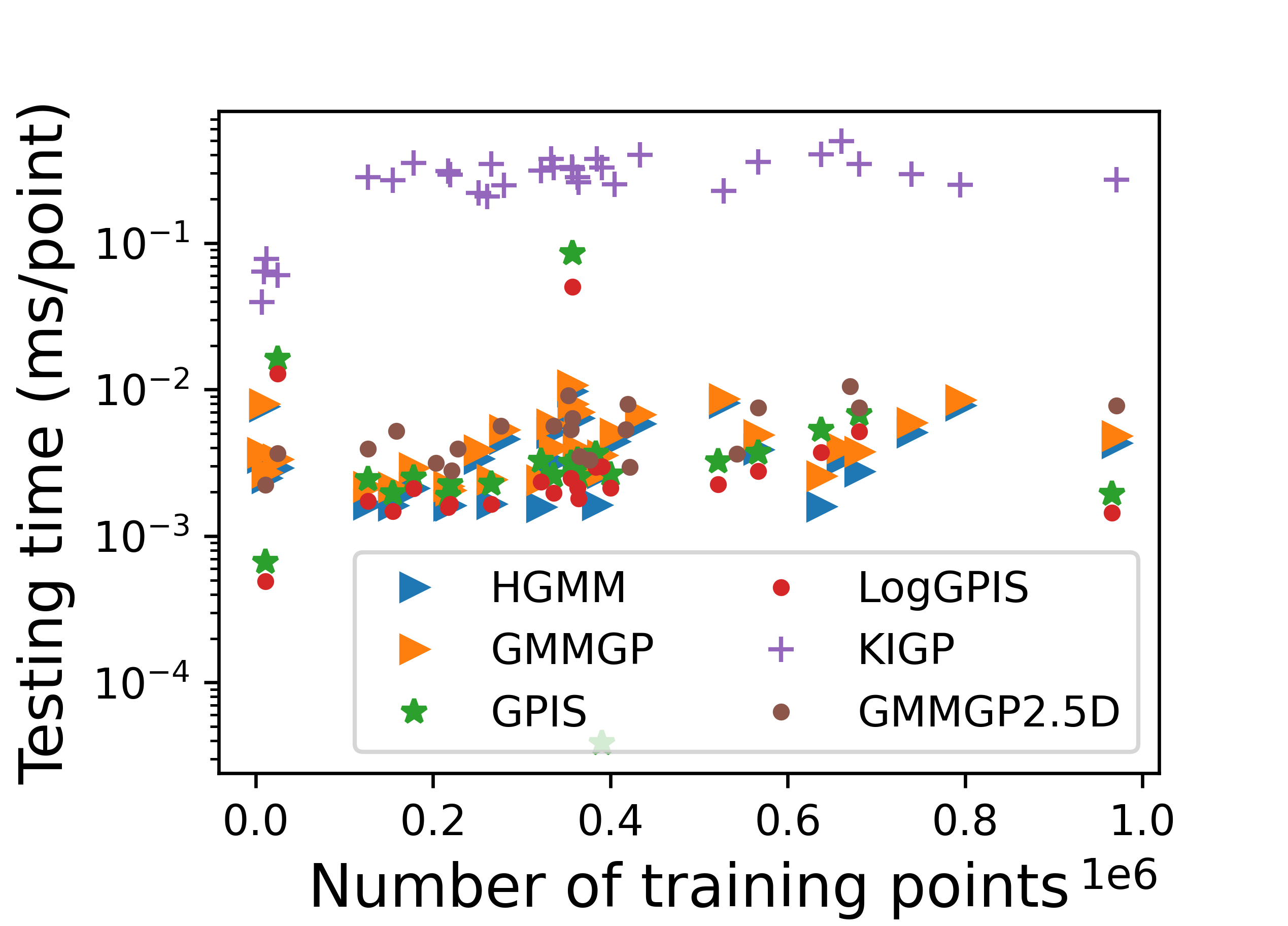}  
  \caption{Testing time}
  \label{fig:test}
\end{subfigure}

\caption{Computational time: note that blue triangles show the HGMM prior estimation while the orange triangles denote the total computation time as the sum of HGMM and GP.}
\label{fig:time_compare}
\end{figure}

Regarding computation time, the proposed GMMGP and HGMM generally reach a faster performance, as shown in Figure \ref{fig:time_compare}. The introduction of the HGMM prior reduces the GP model size in GMMGP, facilitating rapid GP training nearly coinciding with HGMM. Similarly, GMMGP2.5D scales well with large datasets. However, the hierarchical structure and avoidance of segmentation in the proposed GMMGP enhance its efficiency compared to the earlier 2.5D version. During training, the time complexity of GPIS, Log-GPIS and KIGP increases significantly with the number of data points, while the proposed methods remain at a low level. This increase for GPIS and Log-GPIS is primarily attributed to their inherent normal estimation. During the prediction, KIGP incurs high computational costs, whereas performance differences among other methods are less pronounced. Nonetheless, both GPIS and LogGPIS occasionally exhibit significantly higher computational costs, while the proposed methodology maintains a consistently lower and more stable level of time consumption. The HGMM method is often faster, suggesting its viability as a standalone approach for more time-efficient modelling. It is also observed that the total time complexity of GMMGP is greatly affected by the HGMM regression; a more optimized spatial structure for GMR computation will further reduce the computational time. 

\section{Conclusions}

In conclusion, this study has successfully merged the HGMM prior with GP inference to effectively model uncertain implicit surfaces in 3D space. This approach has extended the capabilities of previous models focusing on simple building representations, to encompass more complex structures and surfaces. To harmoniously integrate two probabilistic methodologies, the GMM provides efficient initial surface modelling and furnishes calibrated priors for GP inference. This integration results in a model that is not only compact but also enables the GMMGP to attain more precise estimations of surface distances and generate more reliable uncertainty metrics due to the refined priors. Furthermore, GMMGP exhibits faster computational speed compared to other GP-based frameworks, particularly noticeable during the model's training phase.

Looking ahead, there is potential for further refinement of GMR computations through the adoption of 
spatial index structures.  
Additionally, the method could also allow for the selective use of only accurate estimations in localization and navigation, thereby optimizing performance and utility. The methodology's utility in localization is promising, as the queried surface distances can be directly leveraged for evaluating the alignment quality between two-point clouds in registration.


\section*{Acknowledgments}

The authors acknowledge Bhoram Lee, Lan Wu and Le Gentil for making their code available. 


\vfill

\end{document}